\documentclass[sigconf,noacm]{acmart}
\usepackage{colortbl}
\usepackage{xcolor}
\usepackage{subfiles} 
\usepackage{array}  
\usepackage{multirow}
\usepackage{tabularx}
\usepackage{graphicx} 
\usepackage{makecell}
\usepackage{amsmath}
\usepackage{pifont} 
\usepackage{enumitem}
\usepackage{subfigure}
\usepackage{microtype}
\usepackage{amsmath}
\usepackage{textcomp}

\usepackage{algorithm}
\usepackage{algpseudocode}
\algnewcommand\algorithmicforeach{\textbf{foreach}}
\algdef{S}[FOR]{ForEach}[1]{\algorithmicforeach\ #1\ \algorithmicdo}

\theoremstyle{definition}

\newcommand\vldbdoi{XX.XX/XXX.XX}

\newcommand\vldbavailabilityurl{URL_TO_YOUR_ARTIFACTS}
\newcommand\vldbpagestyle{plain} 

\usepackage[all=normal, floats, bibnotes, wordspacing, charwidths, indent]{savetrees}

\skip\footins 8.1pt plus 4pt minus 2pt 
\newcommand{\revise}[1]{\textcolor{black}{#1}}

\definecolor{grey}{rgb}{0.5,0.5,0.5}
\usepackage{blindtext}

 \setcopyright{none}
\renewcommand\footnotetextcopyrightpermission[1]{} 
\begin{document}

\title{SIMformer: Single-Layer Vanilla Transformer Can Learn Free-Space Trajectory Similarity}

\author{Chuang Yang$^{1}$, Renhe Jiang$^{1}$*, Xiaohang Xu$^{1}$, Chuan Xiao$^{2}$, Kaoru Sezaki$^{1}$}
\thanks{*~Corresponding author.}
\affiliation{%
  \institution{$^{1}$The University of Tokyo, $^{2}$Osaka University}
}
\email{{chuang.yang, jiangrh}@csis.u-tokyo.ac.jp}
\email{xhxu@g.ecc.u-tokyo.ac.jp, chuanx@ist.osaka-u.ac.jp, sezaki@iis.u-tokyo.ac.jp}






\begin{abstract}

Free-space trajectory similarity calculation, e.g., DTW, Hausdorff, and Fr\'e{}chet, often incur quadratic time complexity, thus learning-based methods have been proposed to accelerate the computation. The core idea is to train an encoder to transform trajectories into representation vectors and then compute vector similarity to approximate the ground truth. However, existing methods face dual challenges of effectiveness and efficiency: 1) they all utilize Euclidean distance to compute representation similarity, which leads to the severe curse of dimensionality issue -- reducing the distinguishability among representations and significantly affecting the accuracy of subsequent similarity search tasks; 2) most of them are trained in triplets manner and often necessitate additional information which downgrades the efficiency; 3) previous studies, while emphasizing the scalability in terms of efficiency, overlooked the deterioration of effectiveness when the dataset size grows. To cope with these issues, we propose a simple, yet accurate, fast, scalable model that only uses a single-layer vanilla transformer encoder as the feature extractor and employs tailored representation similarity functions to approximate various ground truth similarity measures. Extensive experiments demonstrate our model significantly mitigates the curse of dimensionality issue and outperforms the state-of-the-arts in effectiveness, efficiency, and scalability.
\end{abstract}

\maketitle

\pagestyle{\vldbpagestyle}


\ifdefempty{\vldbavailabilityurl}{}{
\vspace{.3cm}
\begingroup\small\noindent\raggedright\textbf{Artifact Availability:}\\
The source code, data, and/or other artifacts have been made available at
\url{https://github.com/SUSTC-ChuangYANG/SIMformer}.
\endgroup
}

\section{Introduction}



With the rapid development of positioning and sensing technologies, large-scale trajectory data is being collected from various sources like smartphone apps and navigation systems.
This data holds immense value and plays crucial roles in fields such as intelligent transportation~\cite{wang2018will,fan2019deep}, urban planning~\cite{yuan2012discovering}, and epidemic simulation~\cite{yang2022epimob}.
In particular, \textit{computing the trajectory similarity} is always a fundamental operation for various trajectory analysis tasks, such as clustering~\cite{wang2019fast, agarwal2018subtrajectory}, similarity search~\cite{xie2017distributed, shang2018dita,lan2022vre}, and anomaly detection~\cite{meng2019overview, belhadi2020trajectory}. 
To meet the requirements in different scenarios, many distance functions have been used to calculate the trajectory similarity in free space, such as Dynamic Time Warping (DTW)~\cite{shang2018dita}, Hausdorff distance~\cite{ atev2010clustering}, and Fr\'e{}chet distance~\cite{tang2017efficient}. However, the time complexity of these distance measurements is typically $O(mn)$ (where $m$ and $n$ are the lengths of two trajectories), which limits their application in large-scale datasets~\cite{hu2024spatio,chen2024deep}. 
To this end, a series of approximate approaches to accelerate computation has been proposed, which can be divided into two categories: 
\begin{itemize}[leftmargin=*]
   \item \textit{Non-learning-based methods}~\cite{agarwal2016approximating,ying2016simple,colombe2021approximating,bringmann2016approximability,salvador2007toward} focus on designing more efficient handcrafted approximate algorithms to speed up distance computation. Nevertheless, these methods are often designed for specific one or two distance functions and cannot be easily extended to other trajectory similarity measures.
    \item \textit{Learning-based methods}~\cite{yang2021t3s, yao2019computing, yao2022trajgat, yang2022tmn, zhang2020trajectory,han2021graph, fang2022spatio} 
    concentrate on learning a neural network encoder to transform the original trajectory into a $d$-dimensional representation vector, and then calculate the Euclidean distance between vectors to approximate the target distance measures such as DTW or Hausdorff. This reduces the complexity of trajectory similarity computation to $O(d)$, achieving a shift from quadratic to linear complexity. 
\end{itemize}

Compared to non-learning-based methods, learning-based methods can efficiently perform large-scale similarity calculations on the learned representations. It has been demonstrated that an LSTM-based model can bring a speedup of 50x-1000x over brute-force methods and 3x-500x over non-learning-based methods across three distance measures 
with better top-$k$ hit and recall rates~\cite{yao2019computing}. Besides, learning-based methods can easily approximate different distance measures by simply changing the objective function, showing stronger expandability over non-learning-based ones. Though remarkable progress has been made, similarity learning models under the free-space setting~\cite{yang2021t3s, yao2019computing, yao2022trajgat, yang2022tmn, zhang2020trajectory} still exhibit the following issues.

\begin{figure}[t] 
  \centering
  \includegraphics[width=0.95\linewidth]{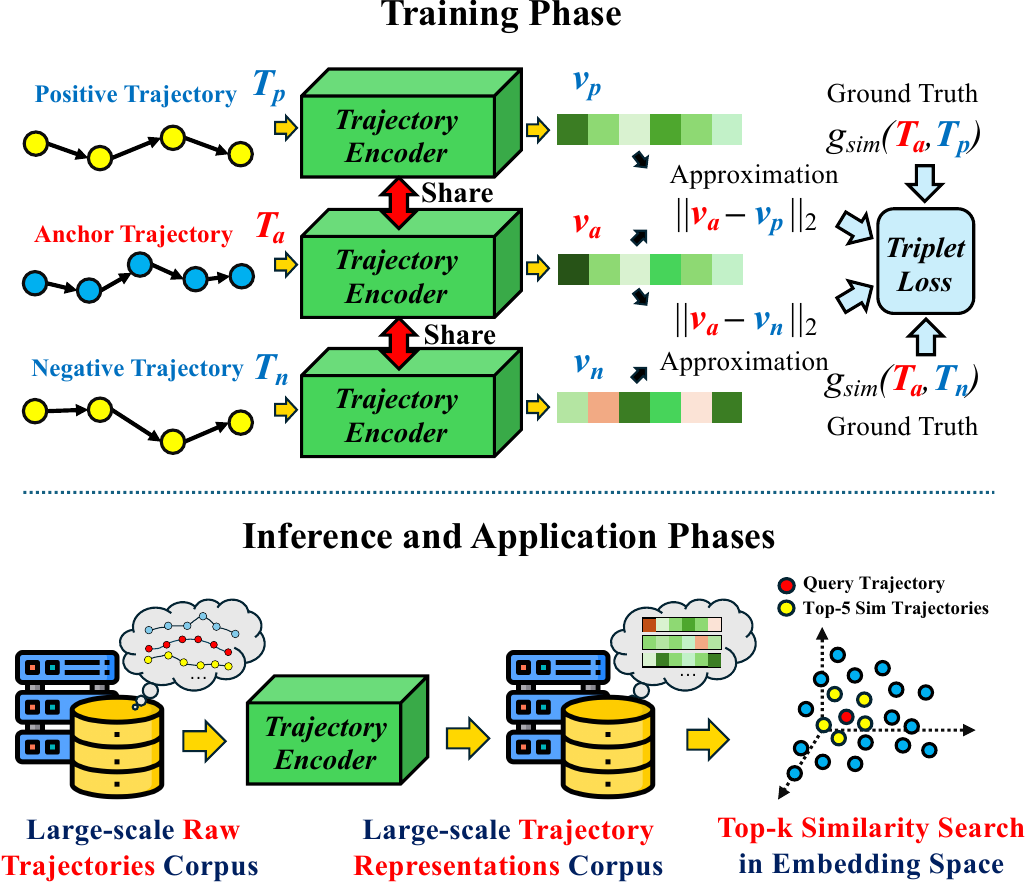}
  \caption{Pipeline of existing learning-based solutions.}
  \label{figure:framework_existing}
\end{figure}

\textbf{(i) Curse of dimensionality and limited effectiveness of learned representations.} Existing studies~\cite{yao2019computing,yao2022trajgat,yang2021t3s,zhang2020trajectory,yang2022tmn} have employed \textit{Euclidean distance-based} representation similarity function to approximate the target measures. In fact, this will lead to severe ``\textit{curse of dimensionality}''~\cite{aggarwal2001surprising} issue, a phenomenon in high-dimensional space such that pairs of objects are not easy to be distinguished, rendering distances, particularly Euclidean distance, to be less effective. 
Such indistinguishability of the latent representations essentially results in poor performance in subsequent tasks, such as the $k$-nearest neighbor ($k$NN) search. Besides, existing works only evaluated the model effectiveness by checking the top-$k$ hit\&recall rate for similarity search. However, \textit{ranking quality}
is also a critical aspect. Overlooking it may lead to high hit ratios but poorly disordered rankings, which inevitably requires re-ranking~\cite{mei2014multimedia}.
\textbf{(ii) Over-complicated models and limited efficiency.}
Following~\cite{yao2019computing}, researchers have adopted the triplet loss that constructs positive\&negative sample pairs for similarity learning as a common technique~\cite{yao2022trajgat, yang2021t3s, zhang2020trajectory, yang2022tmn}, as illustrated in Figure~\ref{figure:framework_existing}. However, this choice greatly increases the training cost. Meanwhile, researchers have suggested employing auxiliary information to improve the performance, like mapping trajectory points onto grids for additional structural insights~\cite{yao2019computing, yao2022trajgat, yang2021t3s} or pre-computing sub-trajectory distances as auxiliary supervision~\cite{zhang2020trajectory, yang2022tmn}.
All these make the state-of-the-art (SOTA) models over-complicated and less efficient. Simplifying the existing framework in a minimalism fashion while making it more efficient and effective becomes the ultimate goal of this study. 
\textbf{(iii) Scalability of effectiveness.} Though scalability is often associated with efficiency, we note significant drops in the effectiveness of current methods as the number of trajectories grows. This is because these methods are typically trained and tested -- also with ground truth labeled -- on a limited portion of the dataset, rendering them less effective against challenging scenarios, such as noise, which predominantly arise in larger datasets. However, this issue has been largely overlooked 
in prior research.

To cope with these limitations, we propose \textit{\textbf{SIMformer}}, a \textit{single-layer vanilla transformer encoder} under a \textit{simple pair-wise mean squared error (MSE) loss} framework, without using the typical triplet loss framework or any auxiliary information. 
In particular, \revise{we find that using \textit{tailored representation similarity functions} (e.g., Chebyshev distance for Hausdorff and cosine for DTW) instead of \textit{relying solely on Euclidean distance} can greatly alleviate the impact of the curse of dimensionality.} This small adjustment allows a solution -- only with a 1-layer vanilla transformer and a Siamese network -- to outperform SOTA in terms of accuracy, speed, and scalability.
\revise{On four widely-used benchmarks Porto~\cite{porto}, T-Drive~\cite{zheng2011t-drive}, Geolife~\cite{zheng2010geolife}, and AIS~\cite{marinecadastre2021}, SIMformer achieves an average improvement of 27.59\% top-$k$ hit ratio on DTW, 34.42\% on Hausdorff, and 12.80\% on Fr\'e{}chet over the best-performing baseline, and improves ranking quality by reducing 20.08\% inversions}. Meanwhile, SIMformer is 30\% faster for inference and saves 10\% memory usage. A scalability test on Porto and its augmented version with random noise shows that SIMformer 
still retains good accuracy when baselines start to report poor results. 
\revise{Comprehensive theoretical and experimental analysis is conducted to reveal and validate how \textit{tailored representation similarity function} alleviates the curse of dimensionality. Furthermore, the applicability of SIMformer across different data and measurement scenarios is explored and discussed.}


Our contributions are summarized as follows:
1) We developed a single-layer vanilla transformer with a simple Siamese architecture to achieve the SOTA performance in both effectiveness and efficiency.
2) To the best of our knowledge, we are the first to study the curse of dimensionality in trajectory similarity learning and propose the idea of \textit{tailoring representation similarity functions} to significantly alleviate this issue.
3) We conducted extensive experiments on \revise{four} widely-used trajectory benchmarks with three distance measures. We also evaluated the \textit{ranking quality} and \textit{scalability} 
to further demonstrate the superiority of the proposed solution. 

\begin{table}[t]
  \centering
  \footnotesize
  \caption{Comparison of model architectures and features.}
  \label{tab:related_work}
  \resizebox{\linewidth}{!}{
  \begin{tabular}{c|c|c|c|cc}
    \hline 
    \multirow{2}{*}{\textbf{Model}} & \multirow{2}{*}{\textbf{\makecell[c]{Core \\ Architecture}}} & \multirow{2}{*}{\textbf{\makecell[c]{Repr. Sim. \\ Func.}}} & \multirow{2}{*}{\textbf{\makecell[c]{Loss \\ Func.}}} &  \multicolumn{2}{c}{\textbf{Need Auxiliary Info.}} \\
    \cline{5-6}
    & & & & \textit{Sub-traj.} & \textit{ Gridification} \\
    \hline
    NeuTraj~\cite{yao2019computing} & Augmented LSTM & Euclidean & Triplet & No & Yes \\
    Traj2simvec~\cite{zhang2020trajectory} & Vanilla LSTM & Euclidean & Triplet & Yes & No \\
    T3S~\cite{yang2021t3s} & LSTM \& Attention & Euclidean & Triplet & No & Yes \\
    TMN~\cite{yang2022tmn} & LSTM \& Attention & Euclidean & Triplet & Yes & No \\ 
    \textbf{SIMformer} & \textbf{Vanilla Transformer} & \textbf{Tailored} & \textbf{MSE} & \textbf{No} & \textbf{No} \\
    \hline 
  \end{tabular}
  }
\end{table}

\section{Related Work}
Existing trajectory similarity learning methods can be divided into two categories based on whether the measure considers the topological structure, which can be built from the proximity relationship between trajectory points~\cite{han2021graph} or the underlying road network~\cite{fang2022spatio}. GTS~\cite{han2021graph} is a graph-based approach proposed for approximate TP distance (an extension of Hausdorff distance on spatial networks). ST2Vec~\cite{fang2022spatio} is built for spatial-temporal trajectory similarity learning in road networks, supporting network-based distance measures like LCRS~\cite{yuan2019distributed} and NetERP~\cite{koide2020fast}.
In comparison, \textit{free-space} trajectory similarity learning, which is not constrained by topological structure, has broader applications and attracts more research attention: NeuTraj~\cite{yao2019computing} is an attention-augmented LSTM model that incorporates spatial context by mapping trajectories into grids and introduces a weighted ranking loss (a triplet loss variant) to improve the performance; Traj2simvec~\cite{zhang2020trajectory} uses a sampling strategy with a k-d tree and $k$NN to accelerate training and sub-trajectory distances for auxiliary supervision; T3S~\cite{yang2021t3s} employs self-attention mechanism with LSTM and further utilizes grid information by incorporating the grid sequence of trajectories as supplementary structural data; 
TMN~\cite{yang2022tmn} uses LSTM and cross-attention for explicitly modeling matching information among trajectories to improve accuracy. 
We outline the model structures and characteristics of these methods in \autoref{tab:related_work}, along with our model. 
It is evident that SIMformer is more concise in design, as it does not require any auxiliary information or the construction of triplets for training.


\section{Preliminaries}
\textit{Definition 1.} (\textbf{Trajectory}): A trajectory $T$ is defined as a time-ordered sequence of locations $\{l_1, l_2, \ldots, l_n\}$, where $n$ is the number of points in the trajectory, and $l_i = (x_i, y_i)$ denotes the $i$-th point in the trajectory. Following existing studies~\cite{yao2019computing,yang2021t3s,zhang2020trajectory,yang2022tmn}, we focus on the shape of trajectories without considering time. 


\noindent \textit{Definition 2.} (\textbf{Free-space Trajectory Distance Measure}):
Given two trajectories $T_i$ and $T_j$, and a free-space trajectory distance measure $M$, $d_{M}(T_i, T_j)$ quantifies the distance between trajectories $T_i$ and $T_j$ without considering the physical constraints in the environment such as the road network.

\noindent \textit{Definition 3.} (\textbf{Ground Truth Trajectory Similarity}): Given two trajectories $T_i$ and $T_j$, the ground truth trajectory similarity $g_{sim}(T_i, T_j)$ is the result of applying negative exponential normalization to trajectory distance $d_M(T_i, T_j)$: 
\begin{equation}
g_{sim}(T_i, T_j) = \exp(-\alpha*d_M(T_i, T_j)) \in [0, 1],
\end{equation}
where $\alpha$ is an adjustable parameter, controlling the distribution of similarity values. This transformation was first introduced in~\cite{yao2019computing} and has been widely adopted in later studies.
It can be treated as a smoothing operation that maps an indefinite range of distance distributions to the range [0, 1], making it easier for models to learn. 

\noindent \textit{Definition 4.} (\textbf{Approximate Similarity Function}): Given two trajectories $T_i$ and $T_j$, the approximate similarity function $f(T_i, T_j)$ approximates the true trajectory similarity $g_{sim}(T_i, T_j)$. It comprises a trainable neural network encoder $f_{enc}$ that maps an input trajectory $T$ to a $d$-dimensional vector $v$, called the trajectory representation:
\begin{equation}
f_{enc}: T \mapsto v \in \mathbb{R}^d,
\end{equation}
and a \textit{representation similarity function} $f_{sim}(\cdot, \cdot)$ that calculates the similarity between two trajectory representations with a time complexity of $\O(d)$. Formally, the approximate similarity function is 
\begin{equation}
\begin{aligned}
    f(T_i, T_j) &= f_{sim}(f_{enc}(T_i), f_{enc}(T_j)), \\ 
                &= f_{sim}(v_i, v_j).
\end{aligned}
\end{equation}

\noindent \textit{Definition 5.} (\textbf{Free-space Trajectory Similarity Learning}): Given a free-space trajectory distance measure $M$, and a training dataset $\mathscr{T}$, the objective is to learn a trajectory encoder $f_{enc}$ such that the discrepancy between the ground truth similarity $g_{sim}(T_i, T_j)$ and the approximated similarity $f(T_i, T_j)$ for trajectory pairs $(T_i, T_j)$ over the training set is minimized, denoted as:
\begin{equation}
\min_{\theta} \sum_{(T_i, T_j) \in \mathscr{T}} \left| g_{sim}(T_i, T_j) - f(T_i, T_j; \theta) \right|, 
\end{equation}
where $\theta$ is the parameters of the encoder $f_{enc}$.
In this study, we select three representative free-space distance measures for ground truth similarity: DTW distance, Hausdorff distance, and Fr\'e{}chet distance.
For Fr\'e{}chet, the discrete version of it is utilized to accommodate the discrete nature of trajectories, inline with previous studies~\cite{yao2019computing,yang2022tmn,yang2021t3s}.


\section{Methodology}

\begin{figure}[t] 
  \centering
  \includegraphics[width=1\linewidth]{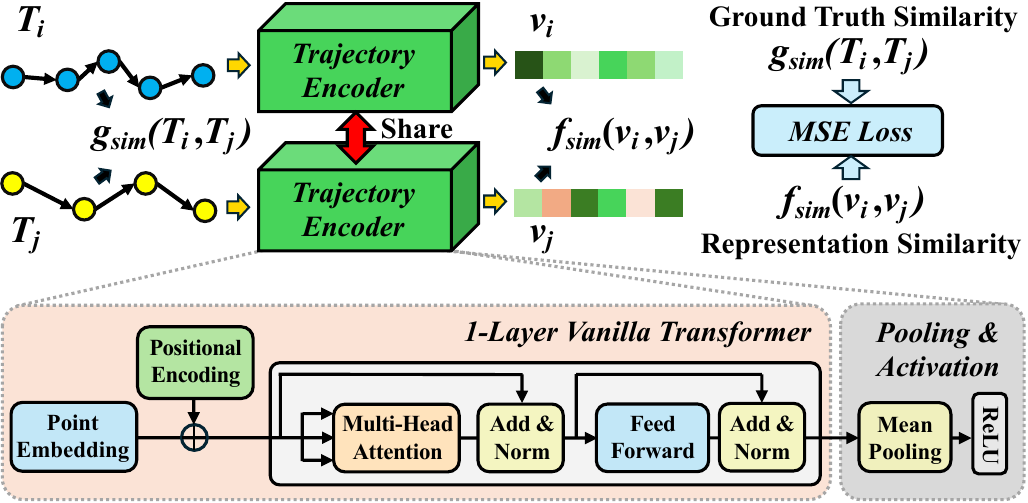}
  \caption{The overall framework of SIMformer.}
  \label{figure:framework_simformer}
\end{figure}

\autoref{figure:framework_simformer} shows the overall framework of the proposed solution, which utilizes a Siamese network architecture.
Given a pair of trajectories $T_i, T_j$, the shared trajectory encoder transforms them into $d$-dimensional representations, and then computes their similarity with the similarity function $f_{sim}(v_i,v_j)$ to approximate the ground truth $g_{sim}(v_i,v_j)$. The Mean Squared Error (MSE) is employed as the training loss. 
We utilize a single-layer vanilla transformer to extract the features, following a pooling \& activation layer to summarize the sequence output, making it length-independent. As for the similarity function, we adopt a tailored approach -- customizing the most appropriate one for different target measures. 
Because of its \underline{\textbf{sim}}plicity and the use of the trans\textbf{\underline{former}} encoder for \underline{\textbf{sim}}ilarity learning, our model is named \textbf{SIMformer}.

\subsection{1-Layer Transformer Trajectory Encoder}\label{sec:encoder}

\noindent \textbf{Point Embedding.}
Each point in the trajectory, denoted as $l_i$, is mapped to $\mathbb{R}^d$ with a linear layer first. The transformation is defined as: $z_i = Wl_i + b$,
where $W \in \mathbb{R}^{d \times 2}$ is the weight matrix, $b \in \mathbb{R}^d$ is the bias vector, and $z_i \in \mathbb{R}^d$ is the transformed point embedding.

\noindent \textbf{Positional Encoding.} 
Before feeding the input into the transformer encoder, a common practice is to add \textit{positional encoding}. 
This ensures the order information of points is integrated, as the transformer encoder processes all input elements in parallel and cannot distinguish their order. Here, we adopt a learnable positional encoding $e_i \in \mathbb{R}^d$ for each position $i$, expressed as:
$h_i = z_i + e_i$.

\noindent \textbf{Transformer Encoder.}
Now, we obtained a preliminary trajectory representation:
$H = [h_1, h_2, \ldots, h_n]^\top \in \mathbb{R}^{n \times d}.$
To learn the complex patterns and dependencies within it, $H$ is further fed into a 1-layer vanilla transformer encoder~\cite{vaswani2017attention}, where the multi-head self-attention extracts and fuses the features of key points, producing a sequence of $d$-dimensional vectors: 
\begin{equation}
\begin{aligned}
    H' & = \text{Transformer Encoder}(H), \\ 
       &= [h_1', h_2', \ldots, h_n']^\top \in \mathbb{R}^{n \times d}.
\end{aligned}
\end{equation}
\noindent \textbf{Pooling and Activation.} 
Lastly, we utilize \textit{mean pooling} to summarize the information within the embedding sequence $H'$, resulting in a $d$-dimensional representation $h_{avg}'$:
$h_{avg}' = \frac{1}{n} \sum_{i=1}^n h_i'.$
Finally, an ReLU activation function is applied to $h_{avg}'$ to confine the final representation within the first quadrant, ensuring that the cosine similarity of the output vectors ranges from 0 to 1:
$v = \text{ReLU}(h_{avg}') \in \mathbb{R}^d.$

The inference complexity of a single-layer vanilla transformer encoder is $O(n^2)$, where $n$ is the length of the trajectory~\cite{vaswani2017attention}.

\subsection{Tailored Representation Similarity Function} \label{sec:tailored_rsf}
\begin{figure}[!t] 
 \captionsetup{labelfont={color=black,bf}}
  \centering
  \includegraphics[width=0.75\linewidth]{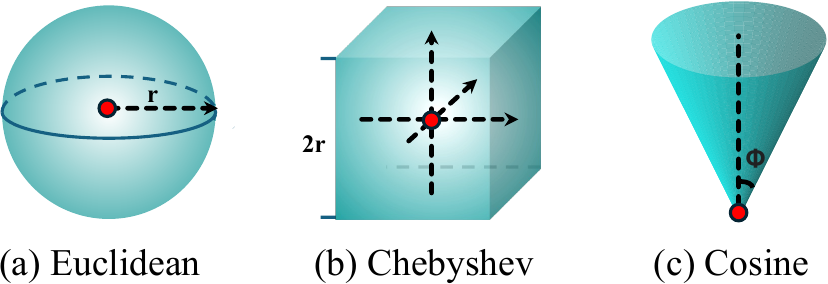}
  \caption{\revise{The feasible solution spaces for different similarity functions in three-dimensional space.}}
  \label{figure:cod_manifold}
\end{figure}

Given two trajectory representation vectors $v_i, v_j \in \mathbb{R}^d$, 
existing learning-based methods~\cite{yao2019computing,zhang2020trajectory,yang2021t3s,yang2022tmn} use the conventional method to compute the representation similarity based on Euclidean distance: 
\begin{equation}
\label{eq:sim_euc}
\begin{aligned}
f_{sim}(v_i, v_j) & = \exp(-d_{\text{Euc}}(v_i, v_j)) 
= \exp\Big(-\sqrt{\sum_{k=1}^{d} (v_{i,k} - v_{j,k})^2} \Big).
\end{aligned}
\end{equation} 
Previous studies~\cite{aggarwal2001surprising,kouiroukidis2011effects} have indicated that Euclidean distance-based measure may lead to the severe curse of dimensionality issues. 
It is essential to explore \textit{alternative similarity functions} that are less sensitive to high-dimensional spaces.
\revise{Here, we consider the alternatives from the \textbf{perspective of feasible solution spaces}}:

\revise{Assuming \(v_i\) is the representation vector of anchor trajectory, and \(v_j\) is the representation vector of the query trajectory, their ground truth distance is $r$. When adopting Euclidean distance for approximation, the deep learning model will force \(v_j\) to lie on the surface of a \(d\)-dimensional \textbf{hyperball} $B^d$ with radius \(r\) centered at \(v_i\). In other words, given \(v_i\), the feasible solution space for \(v_j\) under Euclidean setting is the surface of $B^d$. However, if we replace Euclidean distance with Chebyshev distance: \begin{equation}
\label{eq:cheby}
d_{\text{Cheby}}(v_i, v_j) =  \max_k |v_{i,k} - v_{j,k}|,
\end{equation}
the feasible solution space becomes the surface of a \(d\)-dimensional \textbf{hypercube} \(C^d\) centered at $v_i$ with side length \(2r\), where \(B^d\) is inscribed within \(C^d\). 
Similarly, when using cosine: 
\begin{equation}
\label{eq:sim_cos}
d_{\text{cosine}}(v_i, v_j) = \frac{v_i \cdot v_j}{{\|v_i\|}_2*{\|v_j\|}_2}, \\ 
\end{equation} the feasible solution space is the surface of a \(d\)-dimensional \textbf{hypercone} \(K^d\) with apex at $v_i$ and a fixed opening angle. 
Figure~\ref{figure:cod_manifold} illustrates the shape of the feasible solution spaces for different similarity functions when $d=3$. The red marker represents the anchor point. For Euclidean, all feasible solutions reside on a ball; for Chebyshev, they lie on a cube's surface with a side length of $2r$; under cosine similarity, all solutions reside on a conical surface where all points maintain a constant angle to the anchor point.}

\revise{An interesting and counterintuitive fact is that as $d$ increases, the relative surface area of the hyperball compared to that of the hypercone or hypercube approaches zero. This is quite intuitive for hypercone, because its surface expands infinitely, and so have $A_{\text{hypercone}, d}(\phi) = \infty$. For the hyperball and hypercube, the proof is as follows: the surface area of the hyperball~\cite{li2010concise} with radius $r$ in dimension $d$ is given by:
\begin{equation}
A_{\text{hyperball}, d}(r) = \frac{2 \pi^{d/2}}{\Gamma(d/2)} r^{d-1},
\end{equation}
where $\Gamma$ is the gamma function~\cite{gamma_func}, and the surface area of the hypercube~\cite{hypercube} with side length $2r$ is:
\begin{equation}
A_{\text{hypercube}, d}(2r) = 2d\cdot(2r)^{(d-1)} =d \cdot 2^d \cdot r^{d-1}.
\end{equation}
Then, the ratio $R$ of the surface area of the hyperball to that of the hypercube is:
\begin{equation}
R = \frac{A_{\text{hyperball}, d}(r)}{A_{\text{hypercube}, d}(2r)} = \frac{\frac{2 \pi^{d/2}}{\Gamma(d/2)} r^{d-1}}{d \cdot 2^d \cdot r^{d-1}} = \frac{2 \pi^{d/2}}{\Gamma(d/2) \cdot d \cdot 2^d}.
\end{equation}
Using Stirling's approximation for the Gamma function:
\begin{equation}
R \approx \frac{2 \pi^{d/2}}{ \sqrt{2 \pi \frac{d}{2}} \left( \frac{d/2}{e} \right)^{d/2} \cdot d \cdot 2^d } 
=  \left( \frac{2}{\sqrt{\pi d} \cdot d} \right) \left( \frac{\pi e}{2d} \right)^{d/2}.
\end{equation}
As $\lim_{{d \to \infty}} \left( \frac{\pi e}{2d} \right)^{d/2} = 0$, and $\lim_{{d \to \infty}} \frac{2}{\sqrt{\pi d} \cdot d} = 0$.
Thus, $R \to 0$ when $d \to \infty$. Therefore, we conclude that replacing Euclidean distance with cosine similarity or Chebyshev distance results in a larger potential solution space as dimension increases. Relatively, the solution space for Euclidean contracts into a small area, which is the key source of the curse of dimensionality~\cite{verleysen2005curse}. Experimental validation of this theoretical assertion is presented in Section~\ref{sec:inter}.}

\revise{Despite these advantages, we find that \textbf{there is no one-fits-all solution}.} It is crucial to carefully choose the appropriate similarity function for each distance measure. Specifically, all existing free-space distance measures require finding the optimal matching point for each point~\cite{hu2024spatio,yang2022tmn}. However, the way these matching points are utilized differs significantly: DTW accumulates the distances of all matching point pairs along the optimal warping path, whereas Hausdorff and Fr\'e{}chet distances emphasize the maximum distance among all matching pairs.
Therefore, the similarity function for DTW should accentuate global features, while for Hausdorff and Fr\'e{}chet, it should emphasize critical local differences. 
\revise{Based on this observation, we propose a \textbf{heuristic approach} for further selecting representation similarity functions: when the distance measure considers accumulated global differences (as in DTW), cosine similarity is more suitable because it accounts for global variations, while for measures focused on the best matching pair, such as Hausdorff and Fréchet, using Chebyshev that natively considered only the maximum component difference becomes a more elegant solution.} Such observation is confirmed in our experiments (Table~\ref{tab:top_k_acc}).
As such, we propose to use the ``\textit{tailored representation similarity function}'', whose core idea of is to identify an appropriate similarity based on the specific target measure, rather than rigidly adhering to Euclidean distance. 
\autoref{eq:tailor_all} summarizes the cases for the three distance measures discussed in this paper. 
\begin{equation} \label{eq:tailor_all}
f_{sim}(v_i, v_j) =
\begin{cases}
d_{\text{cosine}}(v_i, v_j) & \text{if } d_{M} = \text{DTW}, \\
\exp(-d_{\text{Cheby}}(v_i, v_j)) & \text{if } d_{M} = \text{Haus./Fr\'e{}c}. \
\end{cases}
\end{equation}

Finally, we employ the MSE loss $\mathcal{L}$ as the optimization objective: 
\begin{equation}
  \mathcal{L} = \frac{1}{N \cdot S}\sum_{i=1}^{N}\sum_{j=1}^{S} \left( f_{sim}(v_i, v_j) - g_{sim}(T_i, T_j) \right)^2,
\end{equation}
where $N$ is the number of trajectories in the training data, and $S$ is the number of trajectories sampled to form input pairs for each $T_i$.

\section{Experiment} \label{sec:exp}
\subsection{Experimental Setup}

\noindent \textbf{Datasets.} 
We utilized \revise{four} trajectory benchmarks, namely Porto~\cite{porto}, \revise{T-Drive~\cite{zheng2011t-drive}}, Geolife~\cite{zheng2010geolife}, and \revise{AIS~\cite{marinecadastre2021}, which encompass taxi trajectories in Porto and Beijing, and individuals' movement trajectories in Beijing, and vessel trajectories in U.S. waters,} respectively. They have been used in previous studies~\cite{yao2019computing,yang2021t3s,yang2022tmn,zhang2020trajectory, hu2024spatio}.
For Porto and Geolife, we conducted the same data preprocessing as in \cite{yao2019computing,yang2022tmn}: trajectories that are too long ($>200$ points), too short ($<10$ points), or located too far from the central city area are removed. \revise{For T-Drive and AIS, which primarily consist of \textit{long-term, continuously tracked trajectories} that have not been segmented into trips, we first segmented them according to stay points~\cite{zheng2015trajectory}, Then, the segmented trajectories were processed following the same procedures as Geolife and Porto. Moreover, the spatial span of the AIS data nearly covers all U.S. waters. Hence, for this study, we only filtered out data for 2021 in Pacific Islands.} 
Considering the high cost of calculating ground truth distance as reported in \cite{yang2021t3s}, we randomly sampled 10,000 trajectories from each dataset for experimentation (except for the scalability test), consistent with previous studies~\cite{yao2019computing,yang2022tmn, yang2021t3s}. For scalability, we also considered an augmented version of Porto with \textbf{1M} trajectories by duplication and inclusion of random noise (Section~\ref{sec:scalability}). 
\autoref{tab:data_preprocessing} summarizes the dataset statistics. 

\begin{table}[t]
  \small
  \centering
   \captionsetup{labelfont={color=black,bf}}
  \caption{\revise{Statistical information of the datasets.}}
  \label{tab:data_preprocessing}
  \resizebox{\linewidth}{!}{
  \begin{tabular}{c|c|c|c|c}
    \hline
    \textbf{Dataset} & \textit{Porto} & \revise{\textit{T-Drive}} & \textit{Geolife} & \revise{\textit{AIS}}\\ \hline
    {Data source} & Taxi & \revise{Taxi} & Human & \revise{Vessel} \\ \hline
    {\#Moving objects} & 442 & \revise{10,357} & 182 & \revise{82,030} \\ \hline
    Length range & [10, 200] & \revise{[10, 200]} & [10, 200] & \revise{[10, 200]} \\ 
    Longitude range & [-9.0, -7.9] & \revise{[115.9, 117.0]} & [115.9, 117.0] & \revise{[-158.51, -157.41]} \\
    Latitude range & [40.7, 41.8] & \revise{[39.6, 40.7]} & [39.6, 40.7]  & \revise{[20.62, 21.72]}  \\ \hline
    \#Trajs. & 599,632 & \revise{15,314} & 11,169 & \revise{10,700} \\ \hline
    \revise{\#Average length} & \revise{49.86} & \revise{69.99} & \revise{76.73} & \revise{80.45} \\ \hline
  \end{tabular}
  }
\end{table}


\noindent \textbf{Baselines.}
We used the following methods as baselines: \textit{NeuTraj}~\cite{yao2019computing} incorporates a spatial attention memory-enhanced LSTM as encoder and distance-weighted triplet loss; \textit{Traj2simvec}~\cite{zhang2020trajectory} uses the vanilla LSTM as encoder and sub-trajectory distances as extra supervision information; \textit{T3S}~\cite{yang2021t3s} combines self-attention and LSTM in an encoder to extract both structured and spatial information; \textit{TMN}~\cite{yang2022tmn} employs LSTM and cross-attention to cope with intra-and inter-trajectory information. Since its output representation is pair-specific and cannot be reused for large-scale computation~\cite{chang2023trajectory}, we used the \textbf{n}on-\textbf{m}atching version suggested by the authors, named TMN-NM.
NeuTraj and TMN's source codes were available. Other baselines were implemented by ourselves. \revise{Additionally, we implemented three ablation variants of SIMformer using different similarity functions: Euclidean, cosine, and Chebyshev.}

\noindent \textbf{Parameter Settings.} 
For each dataset, we divided it into training, validation, and test sets with a ratio of 2:1:7. The hidden dimension $d$ of our model was set to 128, with a learning rate of 0.0005. We used a 1-layer transformer encoder with 16 heads for multi-head self-attention. The batch size for training was set to 20. Following the same sampling number used in most baselines (NeuTraj, T3S, and TMN), each trajectory was randomly paired with 20 other trajectories to construct the training samples, and the traj-dist library~\cite{besse2016review} was used to compute the ground truth.
For $\alpha$, we assigned a value of 16 for DTW, and 8 for both Hausdorff and Fr\'e{}chet distances, consistent with previous studies \cite{yang2022tmn,yao2019computing,yang2021t3s}.
Each experiment was run 3 times, and the average results were reported. 

\noindent \textbf{Environment.} Experiments were conducted on Intel Xeon Silver 4210R CPU @ 2.40GHz and 4 NVidia GeForce RTX 3090 GPUs.

\noindent \textbf{Evaluation Metrics.}
Following existing works~\cite{yang2022tmn,yao2019computing,yang2021t3s,zhang2020trajectory}, we utilize the \textit{top-$k$ hit ratio} (HR@$k$) and \textit{top-$k$ recall for top-$t$ ground truth} (Recall-$t@k$) as the evaluation metrics. Both are used to evaluate the effectiveness of models in top-$k$ similarity search: 
HR@$k$ measures the overlap between the top-$k$ trajectories retrieved using the approximate similarity and those retrieved using the ground truth similarity. Recall-$t@k$ evaluates the model's ability to recall the top-$t$ ground truth items within the retrieved top-$k$ results.
\revise{Moreover, there are several methods for assessing the \textit{ranking quality}, such as Spearman's/Kendall rank correlation coefficient~\cite{rank_cor} and the number of inversions~\cite{inversion}. 
We opted the latter one, 
as it directly reflects the re-ranking cost within the retrieved top-$k$ list.} We compared the order of approximate similarity with the order of ground truth similarity to calculate the inversions. Smaller values are preferred.
\begin{table*}[!htb] 
\centering
\captionsetup{labelfont={color=black,bf}}
\footnotesize
\caption{\revise{Effectiveness in top-$k$ similarity search performance ($\uparrow$)}}
\label{tab:top_k_acc}
\resizebox{\textwidth}{!}{
\begin{tabular}{c|c|cccc|cccc|cccc}
\hline 
\multirow{2}{*}{\textbf{Dataset}} & \multirow{2}{*}{\textbf{Model}} & \multicolumn{4}{c|}{\textbf{Top-$k$ Similarity Search@DTW}} & \multicolumn{4}{c|}{\textbf{Top-$k$ Similarity Search@Hausdorff}} & \multicolumn{4}{c}{\textbf{Top-$k$ Similarity Search@Fr\'e{}chet}} \\
& & \textit{HR@1} & \textit{HR@10} & \textit{HR@50} & \textit{R10@50} & \textit{HR@1} & \textit{HR@10} & \textit{HR@50} & \textit{R10@50} & \textit{HR@1} & \textit{HR@10} & \textit{HR@50} & \textit{R10@50} \\
\hline 
\multirow{8}{*}[0.1cm]{Porto} & NeuTraj~\cite{yao2019computing} & \cellcolor{grey!20}0.2601 & 0.4330 & 0.5540 & 0.7843 & 0.2629 & 0.4043 & 0.5434 & 0.7667 & 0.4056 & 0.5663 & 0.6756 & 0.9324 \\
& Traj2simvec~\cite{zhang2020trajectory} & 0.2599 & \cellcolor{grey!20}0.4540 & \cellcolor{grey!20}0.5826 & \cellcolor{grey!20}0.8244 & 0.3324 & 0.4706 & 0.5893 & 0.8093 & 0.4392 & 0.5999 & 0.6906 & 0.9402 \\
& T3S~\cite{yang2021t3s} & 0.2439 & 0.4305 & 0.5630 & 0.7964 & 0.3493 & 0.5386 & 0.6668 & 0.9226 & \cellcolor{grey!20}0.4623 & 0.6140 & \cellcolor{grey!20}0.7094 & 0.9597 \\
& TMN-NM~\cite{yang2022tmn} & 0.2542 & 0.4440 & 0.5748 & 0.8106 & 0.3905 & 0.5714 & 0.6861 & 0.9427 & 0.4594 & \cellcolor{grey!20}0.6146 & 0.7088 & \cellcolor{grey!20}0.9614 \\
\cline{2-14}

& SIMformer \textit{w/ Euc.} & 0.2580 & 0.4503 & 0.5809 & 0.8205 & \cellcolor{grey!20}0.4394 & \cellcolor{grey!20}0.6137 & \cellcolor{grey!20}0.7009 & \cellcolor{grey!20}0.9595 & 0.4587 & 0.6101 & 0.7017 & 0.9590 \\
& SIMformer \textit{w/ Cos.} & \textbf{0.3685} &\textbf{0.5939} &\textbf{0.7337} & \textbf{0.9490} & 0.4174 & 0.6051 & 0.7020 & 0.9560 & 0.4410 & 0.6164 & 0.7065 & 0.9565 \\
& SIMformer \textit{w/ Cheby.} & 0.2736 & 0.4696 & 0.6001 & 0.8272 & \textbf{0.4990}& \textbf{0.6971} & \textbf{0.7984} & \textbf{0.9811} & \textbf{0.5378} & \textbf{0.7174} & \textbf{0.8170} & \textbf{0.9821} \\
\cline{2-14}
& \textit{Improvement} & 41.68\% & 30.81\% & 25.94\% & 15.11\% & 27.78\% & 22.00\% & 16.37\% & 4.07\% & 16.33\% & 16.73\% & 15.17\% & 2.15\% \\
\hline
\hline
\multirow{8}{*}[0.1cm]{T-Drive}  & NeuTraj~\cite{yao2019computing}  & 0.1018 & 0.2271 & 0.3564 & 0.4931 & 0.1334 & 0.3104 & 0.4857 & 0.6857 & 0.2630 & 0.4494 & 0.5710 & 0.8340 \\ 
& Traj2simvec~\cite{zhang2020trajectory} & \cellcolor{grey!20}0.1188 & \cellcolor{grey!20}0.2659 & \cellcolor{grey!20}0.4117 & \cellcolor{grey!20}0.5801 & 0.1459 & 0.3205 & 0.4888 & 0.6767 & 0.2494 & 0.4300 & 0.5530 & 0.8054 \\ 
& T3S~\cite{yang2021t3s}  & 0.0919 & 0.2258 & 0.3691 & 0.5026 & 0.1935 & 0.4110 & 0.5701 & 0.8214 & 0.3174 & 0.4980 & 0.6132 & 0.8884 \\ 
& TMN-NM~\cite{yang2022tmn} & 0.0983 & 0.2334 & 0.3757 & 0.5248 & 0.2192 & 0.4442 & 0.6041 & 0.8623 & 0.3243 & 0.5085 & 0.6214 & 0.8988 \\ \cline{2-14} 
& SIMformer \textit{w/ Euc.} & 0.1105 & 0.2436 & 0.3833 & 0.5289 & \cellcolor{grey!20}0.3697 & \cellcolor{grey!20}0.5708 & \cellcolor{grey!20}0.6831 & \cellcolor{grey!20}0.9543 & \cellcolor{grey!20}0.3452 & \cellcolor{grey!20}0.5250 & \cellcolor{grey!20}0.6335 & \cellcolor{grey!20}0.9126 \\
& SIMformer \textit{w/ Cos.} & \textbf{0.1988} & \textbf{0.3600} & \textbf{0.5256} & \textbf{0.7394} & 0.3440 & 0.5346 & 0.6415 & 0.9284 & 0.2847 & 0.4519 & 0.5547 & 0.8468 \\
& SIMformer \textit{w/ Cheby.} & 0.1450 & 0.2916 & 0.4372 & 0.5880 & \textbf{0.4366} & \textbf{0.6679} & \textbf{0.7885} & \textbf{0.9804} & \textbf{0.4104} & \textbf{0.6135} & \textbf{0.7322} & \textbf{0.9441} \\
\cline{2-14}
& \textit{Improvement} & 67.33\% & 35.38\% & 27.66\% & 27.46\% & 18.08\% & 17.01\% & 15.44\% & 2.73\% & 18.87\% & 16.85\% & 15.59\% & 3.46\% \\
\hline
\hline
\multirow{8}{*}[0.1cm]{Geolife} & NeuTraj~\cite{yao2019computing}  & 0.1865 & 0.3165 & 0.4319 & 0.6107 & 0.2093 & 0.3558 & 0.5109 & 0.6843 & 0.3201 & 0.5163 & 0.6685 & 0.8587 \\
& Traj2simvec~\cite{zhang2020trajectory} & 0.1898 & \cellcolor{grey!20}0.3394 & \cellcolor{grey!20}0.4733 & \cellcolor{grey!20}0.6649 & 0.2319 & 0.3890 & 0.5361 & 0.7065 & 0.3320 & 0.5308 & 0.6775 & 0.8664 \\
& T3S~\cite{yang2021t3s} & 0.1550 & 0.2876 & 0.4167 & 0.5954 & 0.1834 & 0.3587 & 0.5362 & 0.7130 & 0.2827 & 0.4976 & 0.6680 & 0.8559 \\
& TMN-NM~\cite{yang2022tmn} & 0.1792 & 0.3058 & 0.4233 & 0.6043 & 0.2781 & 0.4838 & 0.6461 & 0.8427 & 0.3351 & 0.5393 & 0.6906 & 0.8871 \\
\cline{2-14}
& SIMformer \textit{w/ Euc.} & \cellcolor{grey!20}0.2001 & 0.3286 & 0.4368 & 0.6271 & \cellcolor{grey!20}0.3502 & \cellcolor{grey!20}0.5678 & \cellcolor{grey!20}0.7047 & \cellcolor{grey!20}0.9047 & \cellcolor{grey!20}0.3519 & \cellcolor{grey!20}0.5646 & \cellcolor{grey!20}0.7058 & \cellcolor{grey!20}0.9062 \\
& SIMformer \textit{w/ Cos.} & \textbf{0.2331} & \textbf{0.3859} & \textbf{0.5203} & \textbf{0.7005} & 0.2991 & 0.4961 & 0.6349 & 0.8385 & 0.3059 & 0.4892 & 0.6293 & 0.8400 \\
& SIMformer \textit{w/ Cheby.} & 0.1873 & 0.3125 & 0.4240 & 0.6329 & \textbf{0.3870} & \textbf{0.6137} & \textbf{0.7628} & \textbf{0.9241} & \textbf{0.3654} & \textbf{0.5823} & \textbf{0.7473} & \textbf{0.9155} \\
\cline{2-14}
& \textit{Improvement} & 16.49\% & 13.70\% & 9.93\% & 5.35\% & 10.51\% & 8.08\% & 8.24\% & 2.14\% & 3.84\% & 3.13\% & 5.88\% & 1.03\% \\
\hline
\hline
\multirow{8}{*}[0.1cm]{AIS}  & NeuTraj~\cite{yao2019computing} & 0.1618 & 0.3377 & 0.4926 & 0.6912 & 0.1514 & 0.3229 & 0.4999 & 0.6614 & 0.2901 & 0.5126 & 0.6581 & 0.8557 \\ 
& Traj2simvec~\cite{zhang2020trajectory}& \cellcolor{grey!20}0.1813 & \cellcolor{grey!20}0.3604 & \cellcolor{grey!20}0.5046 & \cellcolor{grey!20}0.7184 & 0.2225 & 0.4038 & 0.5726 & 0.7240 & 0.3643 & 0.5916 & 0.7150 & 0.9126 \\ 
& T3S~\cite{yang2021t3s} & 0.1487 & 0.3298 & 0.5005 & 0.6901 & 0.1149 & 0.3012 & 0.5154 & 0.6710 & 0.3292 & 0.5664 & 0.7103 & 0.8959 \\ 
& TMN-NM~\cite{yang2022tmn} & 0.1541 & 0.3326 & 0.4951 & 0.6861 & 0.2465 & 0.4799 & 0.6648 & 0.8496 & \cellcolor{grey!20}0.3758 & \cellcolor{grey!20}0.6013 & \cellcolor{grey!20}0.7274 & \cellcolor{grey!20}0.9272 \\ \cline{2-14} 
& SIMformer \textit{w/ Euc.}  & 0.1589 & 0.3308 & 0.4883 & 0.6808 & \cellcolor{grey!20}0.3166 & \cellcolor{grey!20}0.5611 & \cellcolor{grey!20}0.7232 & \cellcolor{grey!20}0.9172 & 0.3730 & 0.5969 & 0.7251 & 0.9249 \\ 
& SIMformer \textit{w/ Cos.} & \textbf{0.2173} & \textbf{0.4215} & \textbf{0.6005} & \textbf{0.7816} & 0.2403 & 0.4531 & 0.6313 & 0.8211 & 0.2773 & 0.4870 & 0.6417 & 0.8521 \\ 
& SIMformer \textit{w/ Cheby.} & 0.1766 & 0.3558 & 0.5126 & 0.7048 & \textbf{0.3558 }& \textbf{0.5968} & \textbf{0.7588} & \textbf{0.9276} & \textbf{0.3952} & \textbf{0.6283} & \textbf{0.7675} & \textbf{0.9326} \\ \cline{2-14} 
& \textit{Improvement}               & 19.83\% & 16.95\% & 19.02\% & 8.80\%  & 12.36\% & 6.36\% & 4.93\% & 1.13\%  & 5.15\%  & 4.49\%  & 5.51\%  & 0.58\%  \\ 
\hline
\end{tabular}}
\end{table*}

\subsection{Effectiveness Evaluation}
\noindent \textbf{Top-$k$ Similarity Search Performance.}
The performance of different models on the top-$k$ similarity search task is displayed in \autoref{tab:top_k_acc}. \revise{The best and second-best results are highlighted in \textbf{bold} and \colorbox{grey!20}{grey}, respectively.} \revise{SIMformer with tailored simlarity function} significantly outperforms other models across all datasets and all three distance measures. Particularly, on the Porto dataset with DTW as the approximation target, its HR@1 surpasses SOTA by 41.68\%. Additionally, on Porto, It achieves 94.90\%, 98.11\%, and 98.21\% R10@50 across the three distance measures, demonstrating its potential for practical applications.
\revise{Furthermore, SIMformer \textit{w/ Euc.} achieves the second-best results for the Hausdorff across all datasets and on parts of the Fréchet. This underscores the superiority of transformers in capturing global key features compared to LSTM, which is used by all baseline models. Nevertheless, comparing the variants of SIMformer reveals that the primary performance enhancement still comes from selecting the most suitable similarity function, as it notably  reduces the concentration effect} (elaborated in Section~\ref{sec:inter}), making the learned representations more discriminative in high-dimensional space.
\revise{Another point worth noting is, all models, including SIMformer, exhibit decreased performance on T-Drive, Geolife and AIS compared to Porto. This is because that the average lengths in these datasets are much longer, while longer trajectories often correspond to more intricate shapes, making them more challenging to model. However, for Geolife and AIS, which are collected in open-space environments - despite AIS having a longer average length, the model performs better on AIS. We attribute this to the fact that Geolife encompasses multiple transportation modes (walking, cycling, driving, etc.), leading to higher complexity in spatiotemporal features~\cite{hu2024spatio}, making it more difficult to learn.}

\noindent \textbf{Ranking Quality.}
\autoref{tab:ranking_error} shows the ranking quality on Porto dataset. SIMformer achieves best performance across all settings. Notably, as $k$ increases, the advantage of our model becomes more pronounced, reaching up to 28.26\% (Porto, Fr\'e{}chet, $k = 100$). This displays the high order consistency between the similarity predicted by SIMformer and the ground truth.
\revise{Additionally, in most cases, all variants of SIMformer demonstrate better ranking quality than baselines. This is because, unlike baseline models that construct triplets (positive \& negative samples) for training, SIMformer randomly samples trajectory pairs and treats them equally with the MSE loss. This strategy allows our models to focus more on learning the general law of the similarity distribution rather than just the extremes, leading to better ranking performance. Moreover, by comparing the three variants, we can find that the tailored similarity function is another key factor in improving ranking performance. It alleviates 
the concentration effect, enabling the model to better capture the real patterns in the data. We validated these conclusions across all datasets; results for other datasets are available in our code repository due to space limitations.}




\begin{table}[!tb] 
\setlength \tabcolsep{1pt}
\captionsetup{labelfont={color=black,bf}}
\centering
\small
\caption{\revise{Effectiveness in ranking error ($\downarrow$) on Porto dataset.}}
\label{tab:ranking_error}
\resizebox{\linewidth}{!}{
\begin{tabular}{c|ccc|ccc|ccc}
\hline 
  \multirow{2}{*}{\textbf{Model}} & \multicolumn{3}{c|}{\textbf{DTW}} & \multicolumn{3}{c|}{\textbf{Hausdorff}} & \multicolumn{3}{c}{\textbf{Fr\'e{}chet}} \\
 &  \textit{k=10} & \textit{k=50} & \textit{k=100} & \textit{k=10} & \textit{k=50} & \textit{k=100} & \textit{k=10} & \textit{k=50} & \textit{k=100} \\
\hline 
  NeuTraj~\cite{yao2019computing} & 15.75  & 399.92 & 1543.90 & 16.12 & 423.98 & 1643.34 & 15.23 & 364.74 & 1357.41 \\
 Traj2simvec~\cite{zhang2020trajectory} & 16.29 & 400.09 & 1528.22 & 14.89 & 391.55 & 1526.35 & 14.35 & 343.87 & 1297.75 \\
 T3S~\cite{yang2021t3s}  & 16.26 & 406.03 & 1552.41 & 15.49 & 376.12 & 1394.67 & 14.39 & 338.65 & \cellcolor{grey!20}1257.39 \\
 TMN-NM~\cite{yang2022tmn} & 16.26 & 401.92 & 1532.77 & 15.03  & 362.67 & 1342.57 & 14.57 & 345.03 & 1279.75 \\
  \cline{1-10}
 SIMformer \textit{w/ Euc.} & 16.07 & 398.30 & 1518.62 & 14.18  & 343.84 & 1289.51 & 14.27 & 344.53 & 1284.90 \\
  SIMformer \textit{w/ Cos.} & \textbf{14.30} & \textbf{326.53} & \textbf{1185.95} & \cellcolor{grey!20}14.14 & \cellcolor{grey!20}342.07 & \cellcolor{grey!20}1283.07 & \cellcolor{grey!20}14.11  & \cellcolor{grey!20}336.33 & 1263.00 \\ 
 SIMformer \textit{w/ Cheby.} & \cellcolor{grey!20}{15.67} & \cellcolor{grey!20}384.22 & \cellcolor{grey!20}1457.39 & \textbf{12.70}  & \textbf{278.79} & \textbf{984.27} & \textbf{12.15} & \textbf{261.61} & \textbf{902.05} \\
 \cline{1-10}
 \emph{Improvement}  & 8.73\%  & 15.01\% & 18.62\% & 10.17\% & 18.50\% & 23.29\% & 13.88\%  & 22.22\% & 28.26\% \\
\hline 
\end{tabular}
}
\end{table}

\subsection{Efficiency Evaluation}
We evaluated the training and inference efficiency of different models. 
Here, ``inference'' refers to the process of converting trajectories into representations. We used Porto with 2,000 trajectories to assess training efficiency and 10,000 trajectories to assess inference efficiency. 
During the training phase, we adhered to the batch size recommendations specified in the original papers of the baseline models. For the inference phase, we set the batch size to 1 for all models to ensure a fair comparison. Each experiment was conducted 10 times, 
and the average values were reported.

The efficiency is displayed in \autoref{tab:computation_efficiency}, including the comparison of model size. SIMformer has the fastest speed and the lowest GPU memory usage among all models during inference. This can be attributed to two factors: 1) the simplicity of our model -- a single-layer transformer encoder; and 2) the inherent parallelism of the self-attention mechanism~\cite{vaswani2017attention}, allowing it to be more efficient than LSTM models that require sequential processing of data~\cite{karita2019comparative}. 
Additionally, our model ranks second in training efficiency, boasting fewer parameters and a shorter training time. Traj2simvec distinguishes itself with the shortest training time, attributed to its unique and efficient sampling strategy, which only necessitated one pair of positive and negative instances per anchor trajectory.
Despite incorporating many advanced techniques, NeuTraj does not significantly increase the parameter count of the LSTM itself, resulting in the least number of parameters. In contrast, T3S 
uses a large trainable look-up table, 
leading to the highest number of parameters.
\begin{table}[!t]
\centering
\caption{Training and inference efficiency on Porto.}
\label{tab:computation_efficiency}
\footnotesize
\setlength \tabcolsep{2pt}
\begin{tabular}{c|c|cc|cc}
 \hline  
 \multirow{2}{*}{\textbf{Model}} & \multirow{2}{*}{\textbf{{\# Params}}} & \multicolumn{2}{c|}{\textbf{Training}} & \multicolumn{2}{c}{\textbf{Inference}} \\ 
  &  & \textit{$t_{epoch}$} & \textit{$t_{total}$} & \textit{GPU Usage} & \textit{$t_{infer}$} \\
\hline 
 NeuTraj~\cite{yao2019computing} & \textbf{0.12M} & 96.39s & 9542.24s & 2197 MiB & 58.370 ms \\
 Traj2simvec~\cite{zhang2020trajectory} & 0.25M & \textbf{5.60s} & \textbf{1233.01s} & 1149 MiB & 4.759 ms \\
 T3S~\cite{yang2021t3s} & 155.15M & 25.96s & 7841.42s & 3395 MiB & 7.533 ms \\
 TMN-NM~\cite{yang2022tmn} & 0.18M & 23.97s & 7455.89s & 1660 MiB & 8.913 ms \\
 \textbf{SIMformer} & \cellcolor{grey!20}{0.16M} & \cellcolor{grey!20}{7.53s} & \cellcolor{grey!20}{3236.66s} & \textbf{1025 MiB} & \textbf{3.419 ms} \\
\hline 
\end{tabular}
\end{table}

\subsection{Scalability Evaluation}
\label{sec:scalability}

\begin{table}[t]
  \centering
  \small 
  \captionsetup{labelfont={color=black,bf}}
  \setlength \tabcolsep{2pt}
  \caption{\revise{Scalability evaluation on Porto.}}
  \label{tab:scalability}
  \resizebox{0.97\linewidth}{!}{
  \begin{tabular}{c|c|ccccc}
    \hline
    \multirow{2}{*}{\textbf{Measures}}    & \multirow{2}{*}{\textbf{Methods}} & \multicolumn{5}{c}{\textbf{Average Query Processing Time (top-50)}} \\ \cline{3-7}
    & &  \textit{1k}     & \textit{5k}     & \textit{10k}    & \textit{200k} & \textit{500k}   \\ \hline
    \multirow{3}{*}{DTW}         & BruteForce & 0.285s & 1.423s & 2.857s & 56.359s & 142.990s \\ \cline{2-7}
    & \revise{Non-learning~\cite{salvador2007toward}} & \revise{0.152s} & \revise{0.716s} & \revise{1.418s} & \revise{28.244s} & \revise{69.314s}      \\  \cline{2-7}
    & \textbf{SIMformer}  & \textbf{0.0181s} & \textbf{0.0188s} & \textbf{0.0195s} & \textbf{0.0221s} & \textbf{0.0226s}       \\ \hline
    \multirow{3}{*}{Hausdorff}   & BruteForce & 0.148s & 0.739s & 1.479s & 28.921s & 75.495s  \\ \cline{2-7}
    & \revise{Non-learning~\cite{taha2015efficient}} & \revise{0.056s} & \revise{0.184s} & \revise{0.360s} & \revise{7.053s} & \revise{17.404s} \\  \cline{2-7}
    & \textbf{SIMformer}  & \textbf{0.0122s} & \textbf{0.0122s} & \textbf{0.0124s} & \textbf{0.0127s} & \textbf{0.0127s}      \\ \hline
    \multirow{3}{*}{Fr\'e{}chet}     & BruteForce & 0.248s & 1.244s & 2.493s & 50.149s & 125.722s \\ \cline{2-7}
    & \revise{Non-learning~\cite{bringmann2016approximability}} & \revise{0.173s} & \revise{0.808s} & \revise{1.534s} & \revise{31.511s} & \revise{78.644s}      \\  \cline{2-7}
    & \textbf{SIMformer}  & \textbf{0.0166s} & \textbf{0.0168s} & \textbf{0.0168s} & \textbf{0.0195s} & \textbf{0.0195s}       \\ \hline \hline 
    

    \multirow{2}{*}{\textbf{Measures}} & \multirow{2}{*}{\textbf{Methods}} & \multicolumn{5}{c}{\textbf{R10@100}} \\ \cline{3-7} 
    & & \textit{10k} & \textit{100k} & \textit{200k} & \textit{500k} & \textit{1M (aug.)} \\ \hline
    \multirow{3}{*}{DTW} & Traj2simvec & 0.8650 & 0.6386 & 0.5625 & 0.4766 & 0.4079 \\ \cline{2-7}
    
    & TMN-NM & 0.8747 & 0.6543 &0.5756 &0.4882 & 0.4215 \\ \cline{2-7}
    & \textbf{SIMformer} & \textbf{0.9645} & \textbf{0.8105} & \textbf{0.7412} & \textbf{0.6563} & \textbf{0.5789} \\ \hline
    \multirow{3}{*}{Hausdorff} & Traj2simvec & 0.8156 & 0.6483 & 0.6191 & 0.5806 & 0.5679 \\ \cline{2-7}
    & TMN-NM & 0.8719 & 0.6902& 0.649 & 0.5979 & 0.5908 \\ \cline{2-7}
    & \textbf{SIMformer} & \textbf{0.9895} & \textbf{0.9209} & \textbf{0.8878} & \textbf{0.8313} & \textbf{0.7990} \\ \hline
    \multirow{3}{*}{Fr\'e{}chet} & Traj2simvec & 0.9247 & 0.8218 & 0.7729 & 0.7079 & 0.6855 \\ \cline{2-7}
    & TMN-NM & 0.9399 & 0.8429 & 0.7994 & 0.7392 & 0.7234 \\ \cline{2-7}
    & \textbf{SIMformer} & \textbf{0.9594} & \textbf{0.8788} & \textbf{0.8377} & \textbf{0.7787} & \textbf{0.7560} \\ \hline
  \end{tabular}
}
\end{table}

\noindent \textbf{Efficiency Scalability.} 
Following~\cite{yao2019computing}, we evaluated the efficiency scalability. We sampled five subsets from Porto, ranging from 1k to 500k. 
The average query processing time (including inference time) for a top-$50$ similarity search is reported in \autoref{tab:scalability}.
We only compared SIMformer with brute-force and \revise{non-learning methods}, as learning-based methods merely differ in inference
time. \revise{The specialized non-learning algorithms for each measure were implemented, including a DTW approximation algorithm~\cite{salvador2007toward} that offers optimal or near-optimal alignments with time and memory complexity of $O(n)$; a linear-time greedy algorithm to approximate the discrete Fréchet~\cite{bringmann2016approximability}; and a near-linear complexity algorithm for computing the Hausdorff distance~\cite{taha2015efficient}.} The results indicate that all methods exhibit linear growth w.r.t. dataset size; but SIMformer demonstrates a significantly more moderate growth rate, \revise{achieving over 1000x speedup than non-learning methods} on the 500k dataset.
\noindent \textbf{Effectiveness Scalability.} 
The scalability of SIMformer was also assessed concerning similarity search accuracy. Five subsets, from Porto, ranging from 10k to 500k, were randomly sampled (excluding the 2k data used for training). The dataset was further augmented to 1M by duplicating it and adding Gaussian noise ($\mu = 0$, $\sigma = 1$, equivalent to 100 meters) to the duplicated trajectory points.
R10@100 performance was assessed using 1,000 queries. We compared SIMformer with Traj2simvec and TMN-NM, the best-performing baselines for DTW and Hausdorff/Fréchet in terms of hit ratio and recall (\autoref{tab:top_k_acc}).
From the scalability results in \autoref{tab:scalability}, the general trend is that all competitors exhibit decreasing effectiveness when the dataset size grows. Nonetheless, SIMformer maintains a hit rate of over 50\% for DTW and over 75\% for the other two distances, even on the 1M dataset augmented with noise. In contrast, the performances of Traj2simvec and TMN-NM deteriorate sharply, with the recalls on DTW drop below 50\% at 500k. 
For Hausdorff, SIMformer's recall on the 500k dataset was even higher than Traj2simvec's on the 10k dataset. 
Moreover, it is evident that the gaps between SIMformer and the baselines are more remarkable on 100k+ datasets across all three distance measures and even exceed 20\% for Hausdorff, showcasing the superior scalability of SIMformer. 
\subsection{\revise{Interpretability Study}}
\label{sec:inter}
\begin{figure}[!t] 
  \centering
  \captionsetup{labelfont={color=black,bf}}
  \includegraphics[width=1\linewidth]{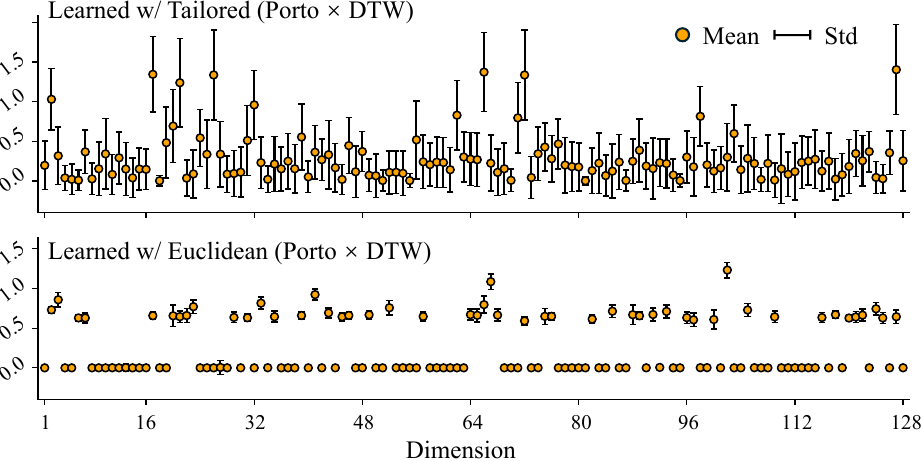}
  \caption{\revise{Distributions of learned representations.}}
  \label{figure:concentration}
\end{figure}

\revise{
According to the theoretical analysis in Section~\ref{sec:tailored_rsf}, Euclidean-based similarity functions are anticipated to have a significantly smaller feasible solution space compared to tailored ones. This assertion is validated here. 
For SIMformer with and without utilizing the tailored similarity function, we respectively calculated the mean and the standard deviation (STD) of learned representations across all 128 dimensions. Figure~\ref{figure:concentration} shows the mean and STD for each dimension when the dataset is Porto and target measure is DTW. It indicates that \textit{minimal variation exists within each dimension when employing the Euclidean-based} similarity function, suggesting that the solutions are confined to a small area. 
Conversely, utilizing the tailored function allows for more flexibility across dimensions, aligning with our theoretical judgement. This phenomenon is called concentration effect~\cite{zimek2012survey} in previous literature, leading to poor performance in downstream tasks.
Notably, this effect is observed consistently across all four datasets $\times$ three distance measures. For other datasets and measures, we also computed the STD of the representations for each dimension. The averaged values are reported (Table~\ref{tab:mstd}), confirming our previous conclusion again.}



\begin{table}[t]
\centering
\captionsetup{labelfont={color=black,bf}}
\footnotesize
\caption{\revise{Average STD of learned representations.}}
\begin{tabular}{l|c|cccc}
\hline
\multirow{2}{*}{\textbf{Measures}}       & \multirow{2}{*}{\textbf{\makecell[c]{Repr. \\ Sim. Func.}}}   & \multicolumn{4}{c}{\textbf{Dataset}} \\ 
& & \textit{Porto} & \textit{T-Drive} & \textit{Geolife} & \textit{AIS} \\ \hline
\multirow{2}{*}{DTW}  & Euclidean  & 0.0315         & 0.0324          & 0.0336           & 0.0674       \\ 
                & Tailored    & \textbf{0.3114}         & \textbf{0.2559}          & \textbf{0.2714}           & \textbf{0.2628}       \\ \hline
\multirow{2}{*}{Hausdorff}   & Euclidean         & 0.0248         & 0.0802          & 0.0478           & 0.0834       \\ 
                & Tailored    & \textbf{0.0647}         & \textbf{0.1517}          & \textbf{0.1034}           & \textbf{0.1481}       \\ \hline
\multirow{2}{*}{Fr\'e{}chet} & Euclidean         & 0.0303         & 0.0903          & 0.0487           & 0.0985       \\ 
                & Tailored    & \textbf{0.0817}         & \textbf{0.2306}          & \textbf{0.1502}           & \textbf{0.2145}       \\ \hline
\end{tabular}
\label{tab:mstd}
\end{table}

\subsection{Hyperparameter Study}
\label{sec:hyper}
\begin{figure}[t]
  \centering
  \includegraphics[width=1\linewidth]{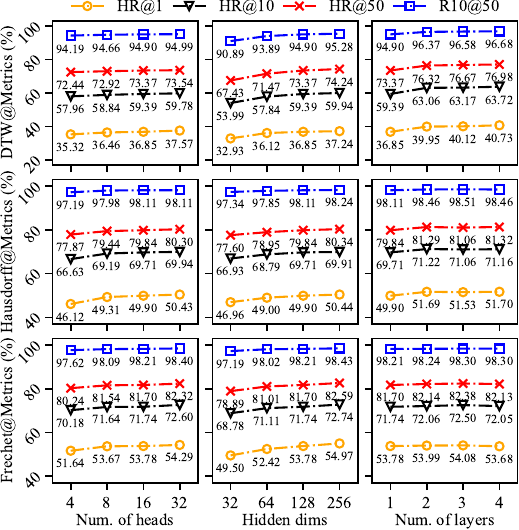}
  \caption{Encoder's hyperparameter analysis on Porto.}
  \label{figure:hyperparameters}
\end{figure}

We tested three hyperparameters of SIMformer's encoder: the number of heads in multi-head self-attention, hidden dimensions of representation, and the number of transformer encoder layers. \autoref{figure:hyperparameters} presents the results on Porto, \revise{from which we can see that by increasing the scale of SIMformer, the model performance can be further improved}. For example, increasing the number of layers from 1 to 4 led to 3\% performance boost across the hit ratios on DTW. 
However, these improvements are accompanied by an increase in model complexity. Considering the model efficiency and simplicity, we opted for moderate parameters: 16 heads, 128 dimensions, and 1 layer for the encoder. 

\section{Discussion and Conclusion} \label{sec:discuss}
\begin{figure}[!t] 
  \centering  \includegraphics[width=0.95\linewidth]{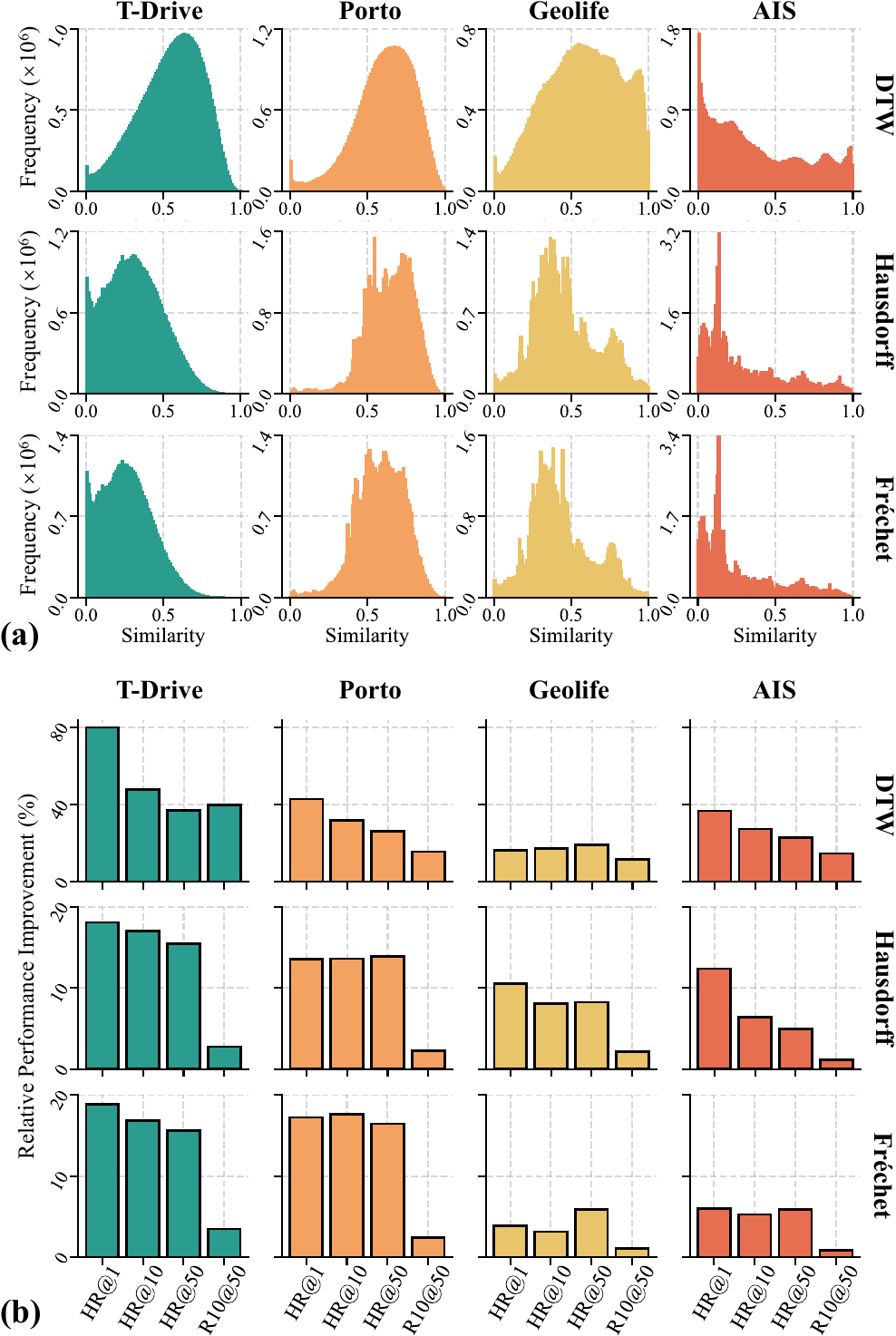}
  \caption{\revise{(a) Ground truth similarity distribution across different datasets and distance measures. (b) Relative performance improvements in top-$k$ query accuracy by replacing Euclidean-based similarity function with the tailored ones.}}
  \label{figure:iid}
\end{figure}

\revise{
\textbf{Applicability vs. Feature Distribution.}
The tailored similarity functions proposed in this paper are designed to mitigate the curse of dimensionality. However, \textit{if the curse of dimensionality is not severe in the data, the improvement from our method may also be limited}. Theoretically, it has been proved that singly distributed data often suffer more from the curse of dimensionality than multiply distributed data because the different distributions inherently provide some separability~\cite{zimek2012survey,beyer1999nearest}. To illustrate this, we plotted the ground truth similarity distributions for various datasets, as shown in Figure \ref{figure:iid}(a). It is evident that the similarity distributions vary significantly across different datasets and distance measures. Overall, the similarity distribution for T-Drive is relatively simple, followed by Porto, with Geolife and AIS becoming increasingly complex. Specifically, Geolife and AIS can be considered multiply distributed data: Geolife includes multiple transportation modes with complex spatiotemporal features, while AIS data are highly unevenly distributed along fixed shipping routes~\cite{hu2024spatio}.}
\revise{We then calculated the relative performance improvement (on top-$k$ query accuracy) achieved by replacing Euclidean distance with tailored similarity functions and visualized these results using bar charts (Figure~\ref{figure:iid}(b)). The results show that \textit{as the data distribution becomes more complex, the performance improvement diminishes}. Consequently, we can reasonably hypothesize that in real-world scenarios, if the spatial distribution of trajectory data is highly uneven (e.g., containing many separated clusters), the performance gains from using tailored similarity functions may be less pronounced.}

\noindent \revise{\textbf{Applicability vs. Distance Measures.} The proposed framework in this paper only addressing two specific cases (\autoref{eq:tailor_all}), leaving another important category of distance measures unaddressed: edit distance-based measures, such as EDR and ERP. 
A key characteristic of these measurements is that their output consists of two heterogeneous components: the \textbf{matched} part and the \textbf{unmatched} part.  In contrast, DTW sums all matches, whereas Fréchet and Hausdorff distances only consider the best one match.
We have conducted some exploratory experiments and found that Euclidean, cosine, and Chebyshev all perform poorly on these \textbf{heterogeneous} distance measurements. Cosine only considers directional similarity, making it difficult to precisely account for both components simultaneously. Chebyshev focuses solely on local optimal information and fails to handle the overall matching situation. While Euclidean distance has an advantage in precision and can capture global information, it still suffers from the curse of dimensionality issue. Therefore, there is a need to ``tailoring another kinds of similarity functions'' that can mitigate the curse of dimensionality while precisely measuring the heterogeneous components.}

\noindent \textbf{Conclusion.} In this study, we proposed a simple yet powerful model for free-space trajectory similarity learning. It utilizes a single-layer vanilla transformer encoder to extract key features from trajectories, along with a tailored representation similarity function for precise similarity approximation towards specific target distance measures, including DTW, Hausdorff, and Fr\'e{}chet. Extensive experiments demonstrated the effectiveness, efficiency, and scalability of our method, as well as the usefulness of the components in the model. We will further investigate tailored similarity functions for more diverse distance measures in the future, with the aim of revealing the overarching design principles. 

\begin{acks}
This work is supported by JST SPRING JPMJSP2108, JSPS KAKENHI Grant Number JP24K02996, JP23K17456, JP23K25157, JP23K28096, and JST CREST Grant Number JPMJCR21M2, JPMJCR22M2.
\end{acks}


\bibliographystyle{ACM-Reference-Format}
\bibliography{sample}



\end{document}